\documentclass[twocolumn]{article} 

\usepackage{multicol}  
\usepackage[letterpaper, top=0.6in, bottom=0.5in, inner=0.75in, outer=0.75in, headsep=0.3in, footskip=0.3in]{geometry}

\usepackage{titlesec}
\titleformat{\section}{\large\bfseries}{\thesection}{1em}{}
\titleformat{\subsection}{\normalsize\bfseries}{\thesubsection}{1em}{}

\usepackage{graphicx}         
\usepackage{amsmath, amssymb} 
\usepackage{hyperref}          
\usepackage{natbib}          
\usepackage{geometry}         
\usepackage{listings}
\usepackage{pgfplots}
\usepackage{amssymb} 
\usepackage{pifont}  
\usepgfplotslibrary{fillbetween}
\usepackage{algorithm}
\usepackage{algorithmic}
\usepackage{amsmath} 
\usepackage{appendix}

\title{\textbf{AlphaSpace: Enabling Robotic Actions through Semantic Tokenization and Symbolic Reasoning}}

\author{
    Alan Dao (Gia Tuan Dao)\textsuperscript{1}, Dinh Bach Vu\textsuperscript{1}, Bui Quang Huy \\
    Menlo Research \\
    \texttt{alan@menlo.ai, bach@menlo.ai, yuuki@menlo.ai} \\
    \textsuperscript{1}Equal contribution.
}

\date{March 25, 2025} 

\begin{document}

\maketitle
\begin{abstract}
\noindent This paper presents AlphaSpace, a novel methodology designed to enhance the spatial reasoning capabilities of language models for robotic manipulation in 3D Cartesian space. AlphaSpace employs a hierarchical semantics-based tokenization strategy that encodes spatial information at both coarse and fine-grained levels. Our approach represents objects with their attributes, positions, and height information through structured tokens, enabling precise spatial reasoning without relying on traditional vision-based embeddings. This approach enables LLMs to accurately manipulate objects by positioning them at specific $(x, y, z)$ coordinates. Experimental results suggest that AlphaSpace demonstrates promising potential for improving manipulation tasks, achieving a total accuracy of 66.67\%, compared to 37.5\% for GPT-4o and 29.17\% for Claude 3.5 Sonnet. These results demonstrate the potential of structured spatial encoding for manipulation tasks and warrant further exploration.
\end{abstract}

\section{Introduction}
\label{sec:introduction}

Large Language Models (LLMs) have demonstrated remarkable capabilities across various domains, including advancements in spatial reasoning \cite{li2024advancingspatialreasoninglarge,wu2024minds}. A significant step in this direction is the AlphaMaze methodology \cite{dao2025alphamazeenhancinglargelanguage} , which introduced a novel approach to enhance LLMs' visual spatial intelligence, specifically for maze navigation. AlphaMaze employs a two-stage training framework, using Supervised Fine-Tuning (SFT) and Group Relative Policy Optimization (GRPO), to enable LLMs to interpret tokenized maze representations and predict step-by-step movement commands. While effective for navigating mazes, this approach presents challenges in larger and more complex environments. This paper introduces "AlphaSpace," an innovative methodology designed to address this challenge and substantially improve the spatial reasoning capabilities of LLMs. "AlphaSpace" builds upon the foundational principles of AlphaMaze by leveraging a semantics tokenization strategy with enhanced semantic tokens to support height (z-coordinate) information.  "AlphaSpace" incorporates synthetic reasoning data, primarily symbolic in nature, to enable the model to move objects to specific given [x, y, z] coordinates \cite{chen2024llmsplanpathsreal,chen2024spatialvlmendowingvisionlanguagemodels,mecattaf2025littleconversationlittleaction}. This includes adding local position information to help reconstruct larger spaces and spawning multiple containers for placing tasks \cite{spatial_reasoning_llm,yamada2024evaluatingspatialunderstandinglarge,daxberger2025mmspatialexploring3dspatial}.

By augmenting the tokenization method with height information and integrating symbolic reasoning data, "AlphaSpace" enables LLMs to understand and manipulate objects within a 3D Cartesian space \cite{wu2025semanticequivalencetokenizationmultimodal, sharma2023exploringimprovingspatialreasoning,xiong20253urllmendtoendmultimodallarge}. This enhancement allows the model to go beyond the 2D limitations and operate in a more complex, three-dimensional environment \cite{chandhok2024scenegptlanguagemodel3d}. Our experiments on embodied manipulation subtasks demonstrate that AlphaSpace achieves 66.67\% total accuracy, significantly outperforming GPT-4o (37.5\%) and Claude 3.5 Sonnet (29.17\%).

\subsection{Motivation}

\label{sec:introduction}
\begin{figure}
    \centering
    \includegraphics[width=1\linewidth]{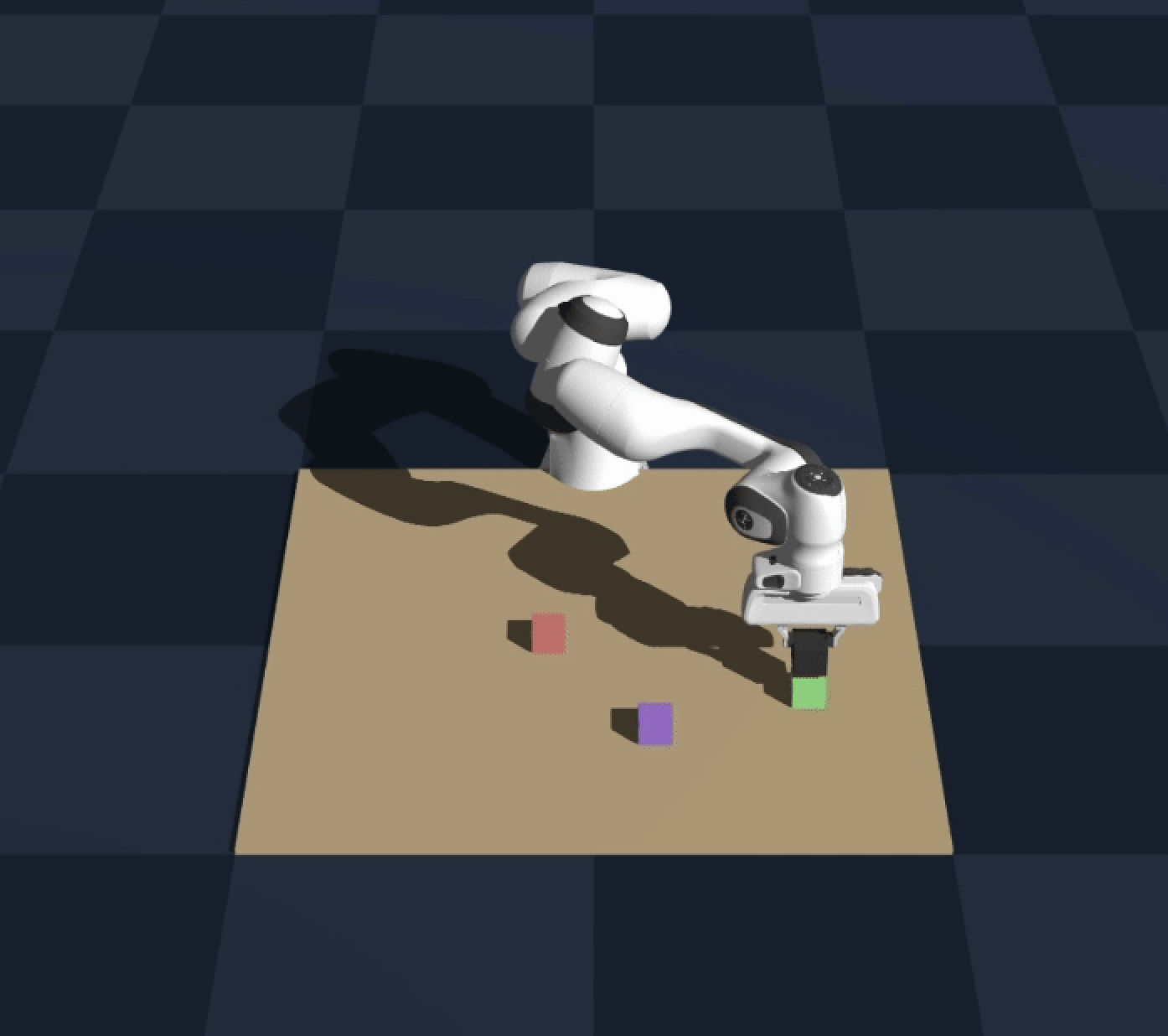}
    \caption{Put black cube onto green cube}
    \label{fig:concept-demo}
\end{figure}
The motivation behind AlphaSpace stems from the current state of robotics models and benchmarks, which primarily rely on either generally pretrained vision-language models (VLMs) \cite{bai2025qwen25vltechnicalreport,yao2024minicpmvgpt4vlevelmllm,hu2024dawnguiagentpreliminary} or research approaches that use vision-language-action (VLAs) \cite{kim2025finetuningvisionlanguageactionmodelsoptimizing, kim2024openvlaopensourcevisionlanguageactionmodel} to predict robotic arm joint angles. However, both of these methods demand extensive training and can be computationally expensive, even during inference.

While generally pretrained vision-language models (VLMs) have demonstrated impressive capabilities in interpreting visual scenes, they struggle with fine-grained spatial reasoning and object manipulation in 3D environments. These models often rely on implicit visual features rather than explicitly structured spatial representations, leading to inconsistencies when performing precise object placement or navigation tasks. Additionally, their performance is constrained by the lack of dedicated spatial priors, making it difficult for them to generalize across diverse manipulation settings. Similarly, approaches that directly predict robotic arm joint angles from visual inputs also faces substantial hurdles. Their reliance on end-to-end learning pipelines often limits adaptability, as minor variations in task conditions can degrade performance. Furthermore, the high-dimensional control space of robotic arms makes it difficult to scale these methods efficiently to more complex tasks.

In contrast, the semantics tokenization strategy has demonstrated that it can provide decoder models with sufficient information to reason about coordinates and spatial attributes in 2D spaces \cite{dao2025alphamazeenhancinglargelanguage}. Many other studies further highlight the ability of decoder-based models to perform spatial reasoning through structured tokenization \cite{wu2024minds,yamada2024evaluatingspatialunderstandinglarge}. By encoding spatial semantics explicitly, these approaches enable models to achieve more structured and interpretable spatial understanding.

Building on these insights, AlphaSpace introduces an advanced tokenization method that extends spatial reasoning beyond two dimensions into a full 3D Cartesian space. By integrating symbolic reasoning data and enriched semantic tokens, AlphaSpace enables more efficient and precise manipulation of objects in three-dimensional environments. Unlike VLAs or VLMs, which heavily depend on the quality of modality encoders—whether for vision \cite{zhai2023sigmoidlosslanguageimage} or actions \cite{pertsch2025fastefficientactiontokenization}—and require intensive computation, AlphaSpace offers a lightweight yet effective approach to spatial reasoning. This structured representation significantly enhances the model’s ability to generalize across tasks, making it particularly well-suited for robotics, object manipulation, and large-scale spatial navigation.

\subsection{Contributions}

Our key contributions are as follows:

\begin{itemize}
    \item Evidence that Decoder-Only Models Can Perform 3D Spatial Reasoning Using Semantic Tokens – Demonstrating that a decoder-only architecture, without reliance on explicit 3D geometric encoders or vision modules, can effectively reason in a 3D Cartesian space using an enhanced semantic tokenization approach. This extends previous 2D spatial reasoning capabilities to support height (z-coordinates) and object manipulation.
    \item Enhanced Semantic Tokenization for 3D Spatial Reasoning – extending the semantics tokenization approach to support 3D Cartesian space, allowing models to reason about height (z-coordinates) in addition to 2D positional attributes.
    \item Symbolic Reasoning for Object Manipulation – Incorporating synthetic reasoning data to enable LLMs to manipulate objects in a structured manner, facilitating precise object placement and spatial transformations.
    \item Empirical Validation on Embodied Manipulation Tasks – Demonstrating AlphaSpace’s effectiveness through experiments on embodied spatial reasoning tasks, where it significantly outperforms existing models such as GPT-4o and Claude 3.5 Sonnet in 3D object manipulation accuracy.
\end{itemize}
\section{Related Work}
\label{sec:related_work}

\subsection{Advancements in Vision-Language-Action Models}

Recent developments in VLA models have significantly enhanced robotic manipulation capabilities. \cite{kim2024openvlaopensourcevisionlanguageactionmodel} introduced OpenVLA, a open-source 7-billion-parameter open-source VLA model trained on 970,000 robot episodes from the Open X-Embodiment dataset. OpenVLA supports controlling multiple robots out-of-the-box and can be quickly adapted to new robot setups via parameter-efficient fine-tuning, setting a new state of the art for generalist robot manipulation policies.

\subsection{General-Purpose Robot Foundation Models}

The \( \pi_0 \)  (pi-zero) model, proposed by \cite{black2024pi0visionlanguageactionflowmodel}, represents a significant advancement in general-purpose robot foundation models. Built on a flow matching architecture atop a pre-trained vision-language model, \( \pi_0 \) is trained on diverse datasets from multiple dexterous robot platforms, including single-arm robots, dual-arm robots, and mobile manipulators. The model demonstrates capabilities in performing zero-shot tasks, following language instructions, and acquiring new skills via fine-tuning, covering a wide array of tasks such as laundry folding, table cleaning, and assembling boxes.

\subsection{Integrating Spatial Intelligence into AI and Robotics}

Several notable developments have contributed to integrating spatial intelligence into AI and robotics. \cite{deepmind2025geminirobotics} introduced Gemini Robotics and Gemini Robotics-ER, models that leverage the reasoning abilities of LLMs to assist robots in performing complex tasks, marking a significant shift in using AI to improve adaptability and performance in various environments.

These advancements collectively contribute to the enhancement of spatial reasoning in LLMs, providing valuable insights and methodologies that align with the objectives of "AlphaSpace" in improving 3D spatial understanding and manipulation capabilities.

\section{Methodology}
\label{sec:methodology}

To develop AlphaSpace, we designed a training pipeline inspired by AlphaMaze to enhance manipulation capabilities within a tabletop environment, specifically targeting the EmbodiedBench benchmark. The methodology consists of a two-stage synthetic dataset generation process, followed by model training utilizing structured tokenization and reasoning-based learning.

\subsection{Synthetic Dataset Generation}

Our synthetic dataset consists of two connected components: object configurations and corresponding action plans.

\subsubsection*{Object Configurations}

The environment consists of a planar tabletop workspace discretized into a $100 \times 100$ grid. To facilitate hierarchical reasoning, this grid is further subdivided into a coarser $25 \times 25$ grid, where each coarse cell encompasses a $4 \times 4$ subgrid of the finer resolution. An object's location is specified by a tuple $(r_g, c_g, r_l, c_l)$, where $(r_g, c_g)$ represent the row and column indices in the coarse grid ($0 \leq r_g, c_g < 25$), and $(r_l, c_l)$ represent the row and column indices within the fine grid of the corresponding coarse cell ($0 \leq r_l, c_l < 4$).

Objects are represented by a tuple $(color, shape)$, where $color$ is selected from a set of 19 distinct colors (red, blue, green, purple, etc.), and $shape$ is one of cube, cylinder, triangular prism, star, moon, or container. During object placement, we enforce constraints to ensure non-overlapping objects (minimum 4-unit Euclidean distance between centers), uniform spatial distribution of objects, and variable object heights (uniformly sampled from [1, 30]).

The dataset encompasses three primary task categories: (1) placement tasks, where the robot must pick up an object and place it into a container (e.g., ``Pick up the red cube and place it into the blue container''); (2) stacking tasks, where one object must be positioned on top of another (e.g., ``Stack the red cube on top of the blue cylinder''); and (3) movement tasks, where objects are relocated to specific coordinates (e.g., ``Move the green star to [76, 65]''). Scene complexity ranges from 4-7 objects per environment, with each scene containing the necessary objects for task completion plus additional objects for environmental complexity.

\subsubsection*{Action Plan Generation}

For each object configuration, we generate a corresponding action plan representing a sequence of robot actions required to achieve the specified manipulation goal. Each action is encoded as a 7-dimensional vector consisting of: global position in the $25 \times 25$ grid $(r_g, c_g)$, local position in the $4 \times 4$ subgrid $(r_l, c_l)$, Z-axis height (vertical distance from gripper to table surface), roll orientation (range: 0-120, each unit representing 3 degrees), pitch orientation (range: 0-120), yaw orientation (range: 0-120), and gripper state (0 for closed, 1 for open).

A typical action sequence progresses through seven steps: (1) approaching above the source object with open gripper, (2) lowering to the object's base, (3) closing the gripper to grasp the object, (4) lifting to a safe height, (5) moving above the target location, (6) lowering to the placement position, and (7) opening the gripper to release the object. Each sequence is accompanied by explicit reasoning annotations that detail the spatial problem-solving strategy, including object localization (identification of relevant objects' positions and properties) and action planning (step-by-step breakdown of the manipulation process).

These reasoning annotations provide valuable supervision signals for training models to understand the connection between perception, spatial reasoning, and action planning. Our final dataset consists of approximately 260,000 synthetic samples, with 100,000 placement tasks, 120,000 stacking tasks, and 40,000 movement tasks. For a subset of the data, we employed object uniqueness constraints to prevent the model from relying on spurious correlations between object types and actions.

\subsection{Model Training Pipeline}

Following dataset generation, we trained the model using a decoder-only architecture with an enhanced semantic tokenization strategy. By integrating enhanced spatial tokenization and structured reasoning data through Supervised Fine-Tuning (SFT) on the synthetic reasoning dataset, AlphaSpace effectively learns spatial relationships and object manipulation tasks, outperforming baseline models in EmbodiedBench evaluations.

\subsection{Evaluation}

We use EmbodiedBench \cite{yang2025embodiedbenchcomprehensivebenchmarkingmultimodal}, a widely used benchmarking framework for evaluating the performance of vision-language models (VLMs) on various vision and spatial reasoning tasks.

Specifically, we focus on the EB-Manipulation subset, which assesses VLMs on spatially related pick-and-place tasks. Within this benchmark, we use a set of 12 pick tasks and 12 place tasks, selected from the spatial tasks category of EB-Manipulation. Hence, the benchmark used in this paper is a subset of EB-Manipulation, specifically focusing on spatial pick-and-place tasks, and comprises a total of 24 actions.

Additionally, we re-ran the benchmark for Claude-3.5-Sonnet and GPT-4o, the two leading models on the current benchmark leaderboard.
\section{Experiments and Results}
\label{sec:experiments}
To evaluate the effectiveness of AlphaSpace in manipulation tasks, we conducted experiments on the EmbodiedBench benchmark, specifically focusing on the Manipulation Subtask.

\subsection{Training Details}

We trained our spatial reasoning model using supervised fine-tuning on the synthetic dataset described in the previous section. Our model is based on DeepSeek-R1-distil-Qwen-1.5B. The training process used a learning rate of $1.0 \times 10^{-4}$ with a cosine decay schedule and a warm-up ratio of 0.1.  Training was conducted on 8 NVIDIA H200 GPUs, with a batch size of 16 samples per device and a maximum context length of 4096 tokens.

\subsection{Evaluation Results}

We evaluated AlphaSpace against state-of-the-art models, including GPT-4o and Claude 3.5 Sonnet. The results on the EmbodiedBench Manipulation Subtask are presented in Table \ref{tab:results} and visualized in Figure \ref{fig:performance}.

\begin{table}[h]
\centering
\begin{tabular}{|c|c|c|c|}
\hline
Model & Picking & Stacking & Total (\%) \\
\hline
AlphaSpace (Ours) & \textbf{10/12} & \textbf{6/12} & \textbf{66.67\%} \\
GPT-4o & 6/12 & 3/12 & 37.5\% \\
Claude 3.5 Sonnet & 5/12 & 2/12 & 29.17\% \\
\hline
\end{tabular}
\caption{Evaluation Results on EmbodiedBench Manipulation Subtask}
\label{tab:results}
\end{table}

As shown in the results, AlphaSpace significantly outperforms both GPT-4o and Claude 3.5 Sonnet across the picking and stacking tasks. Our model achieves 66.67\% overall accuracy, compared to 37.5\% for GPT-4o and 29.17\% for Claude 3.5 Sonnet. This performance gap demonstrates the effectiveness of our approach in spatial reasoning for robotic manipulation tasks.

The significant improvement over commercial state-of-the-art models highlights the advantages of our specialized training approach, which combines synthetic data generation with explicit reasoning supervision. In particular, AlphaSpace demonstrates stronger performance on stacking tasks (50\% success rate) compared to picking tasks (83.3\% success rate), suggesting that compositional manipulation tasks remain challenging even for our specialized model. These results indicate that our approach provides a promising direction for embodied AI systems that require sophisticated spatial reasoning capabilities.

\begin{center}
    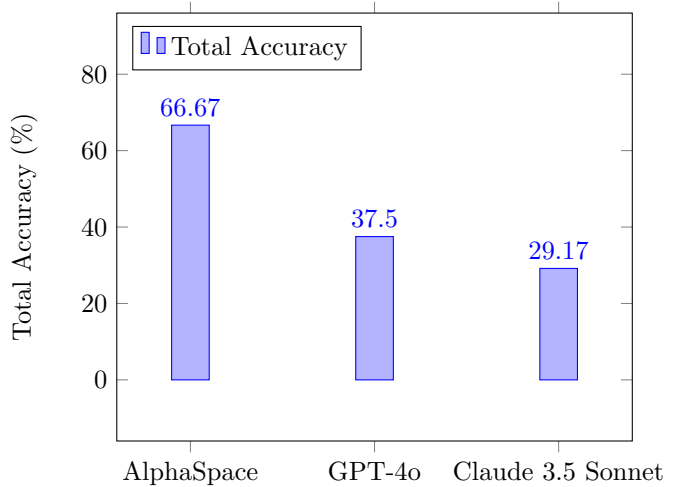
\begin{figure}[h]
\centering
\begin{tikzpicture}
\begin{axis}[
ybar,
symbolic x coords={AlphaSpace, GPT-4o, Claude 3.5 Sonnet},
xtick=data,
ymin=0, ymax=80,
ylabel={Total Accuracy (\%)},
nodes near coords,
bar width=0.5cm,
enlargelimits=0.2,
legend pos=north west
]
\addplot coordinates {(AlphaSpace,66.67) (GPT-4o,37.5) (Claude 3.5 Sonnet,29.17)};
\legend{Total Accuracy}
\end{axis}
\end{tikzpicture}
\caption{Performance Comparison on EmbodiedBench Manipulation Subtask}
\label{fig:performance}
\end{figure}
\end{center}
\section{Discussion}
\label{sec:discussion}
The experimental results demonstrate that AlphaSpace significantly improves the spatial reasoning and manipulation capabilities of LLMs in a 3D Cartesian space. This enhancement is primarily attributed to the novel semantics-based tokenization strategy and the integration of symbolic reasoning data. In this section, we analyze the implications of these results, discuss the strengths and limitations of the approach, and outline potential future directions for research.

\subsection{Strengths}
\label{subsec:strengths}
One of the key advantages of AlphaSpace is its ability to encode height information through semantic tokens, allowing LLMs to reason about three-dimensional spatial structures without relying on traditional vision-based embeddings. Unlike VLMs and VLA models, which extract spatial information from visual encoders such as SigLIP, AlphaSpace directly represents spatial semantics through structured tokenization. This removes the need for intermediate visual interpretation and allows for more precise and computationally efficient reasoning about object relationships. As a result, AlphaSpace serves as a lightweight yet effective alternative for robotic manipulation tasks.

Furthermore, AlphaSpace's reliance on structured symbolic reasoning data enhances its generalizability. The model's ability to infer correct object placements and transformations in novel spatial arrangements suggests that it is learning a more structured form of spatial reasoning compared to end-to-end visual learning approaches. This is evident in its performance on embodied manipulation tasks, where it achieved a 66.67\% accuracy—significantly outperforming state-of-the-art models like GPT-4o and Claude 3.5 Sonnet.

\subsection{Limitations}
\label{subsec:limitations}

Despite its strong performance, AlphaSpace has several limitations. First, the model's reliance on tokenized spatial representations means it may struggle with highly dynamic environments where real-time sensory feedback is crucial. Unlike VLMs, which continuously process visual input, AlphaSpace depends on pre-tokenized spatial descriptions, making it less adaptive to rapidly changing scenarios.

Another limitation is the assumption of a structured environment. The experimental setup ensures controlled object placements and clear spatial references, but real-world applications often involve occlusions, complex object geometries, and unpredictable external forces. Extending AlphaSpace to handle such uncertainties would require either a hybrid approach that integrates limited vision-based feedback or an expansion of the tokenization framework to incorporate uncertainty modeling.

Additionally, while AlphaSpace demonstrates strong performance in tabletop object manipulation tasks, its scalability to more complex robotic scenarios, such as multi-step assembly or dynamic obstacle avoidance, remains an open question. Future work should explore ways to extend its reasoning capabilities beyond static manipulation tasks.

Furthermore, the model does not leverage recent advancements in reinforcement learning for large language models (LLMs), particularly the methods introduced by DeepSeek \cite{shao2024deepseekmathpushinglimitsmathematical, deepseekai2025deepseekr1incentivizingreasoningcapability}. Despite these techniques being proven effective in improving the accuracy of decoder models for solving spatial reasoning tasks with visual components \cite{dao2025alphamazeenhancinglargelanguage}, they have not been incorporated into the current approach.

The last limitation stems from the premature conclusion of the project due to a change in direction. Originally, the study aimed to cover four tasks to fully evaluate the Spatial section of EB-Manipulation, but only two tasks were completed. This restricted the scope of evaluation, potentially leaving gaps in understanding AlphaSpace’s full capabilities. Future iterations should ensure comprehensive task coverage to provide a more complete assessment of its effectiveness.

\subsection{Future Directions}
\label{subsec:future_directions}

To further enhance AlphaSpace, several research directions can be explored. One promising avenue is the integration of reinforcement learning-based fine-tuning to enable real-time decision-making. While AlphaSpace currently relies on a fixed set of symbolic reasoning data, incorporating reinforcement learning could help it adapt to unexpected scenarios more effectively.

Another area of interest is hybrid modeling \cite{noh2021hvprhybridvoxelpointrepresentation}, where AlphaSpace could be combined with lightweight vision modules to bridge the gap between symbolic reasoning and real-world perception. By integrating minimal visual priors, AlphaSpace could potentially enhance its adaptability without incurring the high computational costs associated with full VLM-based architectures.

Finally, expanding the tokenization approach to support dynamic spatial transformations, such as rotating or deforming objects, would be a crucial next step. Current tokenization strategies primarily capture static positions, but real-world robotic tasks often involve complex spatial operations that require continuous adjustment.
\section{Conclusion}
\label{sec:conclusion}

AlphaSpace introduces a novel semantic tokenization strategy and symbolic reasoning approach to enhance 3D spatial reasoning in LLMs. By encoding height information and leveraging structured synthetic reasoning data, AlphaSpace enables accurate object manipulation in Cartesian space. We demonstrate a significant performance improvement over existing models with our experimental results, highlighting the effectiveness of our approach. This work paves the way for more structured, efficient spatial reasoning in robotics and AI, reducing reliance on vision-based encoders and high-dimensional control models. Future work will explore real-world deployment and integration with multimodal systems.

\bibliographystyle{plainnat} 
\bibliography{bibliography} 



\end{document}